\documentclass[10pt,twocolumn,letterpaper]{article}

\usepackage[pagenumbers]{cvpr} 

\usepackage{amsmath}

\definecolor{cvprblue}{rgb}{0.21,0.49,0.74}
\usepackage[pagebackref,breaklinks,colorlinks,allcolors=cvprblue]{hyperref}
\usepackage{float}

\def\confName{CVPR}
\def\confYear{2027}

\title{Consistent Identity Association for Open-Vocabulary Counting in Videos}

\author{Joana Konadu Owusu, Shivanand Venkanna Sheshappanavar\\
Geometric Intelligence Research Lab, Dept. of Electrical Engineering and Computer Science\\
University of Wyoming, 1000 E University Ave, Laramie, WY 82071\\
{\tt\small \{jowusu1, ssheshap\}@uwyo.edu}
}

\begin{document}
\maketitle
\begin{abstract}
Open-vocabulary video object counting asks how many distinct instances of a text-specified category appear across a video. Recent approaches combine promptable detection with tracking and count the resulting identities in the video. However, occlusions, re-entry, and fragmented trajectories can cause trackers to issue new identities to objects already counted. To address this issue, we present Consistent Identity Association (CIA), a lightweight association module that decouples identity counting from tracker outputs. Whenever the underlying tracker issues new identities, CIA jointly determines whether the corresponding detections represent new objects, returning instances, or duplicates of active tracks. This consistent identity assignment reduces repeated counting. We also introduce the Video Counting Benchmark (VCB), comprising 951 videos and 1,639 queries, with instance identities for evaluating whether objects are counted once. We instantiate CIA with a frozen SAM~3 tracker and evaluate it on VCB and VideoCount. On BURST-Count, Dance-Count and GMOT-Count, CIA reduces count MAE by \textbf{60\%}, \textbf{63\%} and \textbf{13\%} relative to CountVid, and by \textbf{3\%}, \textbf{68\%} and \textbf{19\%} relative to SAM3Count, respectively.  Code is available at \href{https://github.com/Joan947/CIA}{https://github.com/Joan947/CIA}.
\end{abstract}
    
\section{Introduction}
\label{sec:intro}

\begin{figure}[!t]
\centering
\includegraphics[width=\columnwidth]{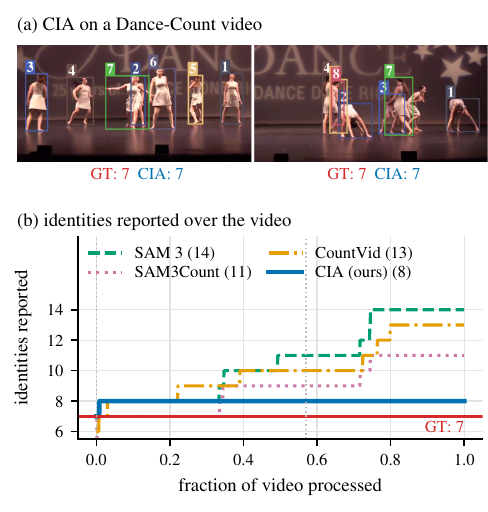}
\caption{Open-vocabulary counting in videos. (a) Two frames of a selected Dance-Count video with 7 labeled dancers and the identities CIA reports. (b) The number of distinct identities each system reports as the video is processed.}
\label{fig:teaser}
\end{figure}

Video object counting determines how many distinct instances of a queried object appear in a video~\cite{owusu2026objectcountingmodalitiestaxonomies}. Applications include monitoring animal populations~\cite{cui2025fishSurvey}, cell and scientific imaging~\cite{TJ_2023_ACCT}, crowd management~\cite{Zhang_2016_Single}, traffic analysis~\cite{dai2019vehicleCounting,NC_2022_Optimal}, and drone surveillance~\cite{wen2021dronecrowd}. Unlike image- or frame-level counting, which estimates the number of visible objects in a single image, video counting must distinguish newly observed objects from those already counted in previous frames.
Tracking-based counting addresses this challenge by associating observations over time to count distinct identities~\cite{amini2025countvid,li2023sceneaware,dai2019vehicleCounting,NC_2022_Optimal}. Multi-object tracking (MOT)~\cite{weng2020gnn3dmot,pang2021qdtrack,zhang2022bytetrack} and Video instance segmentation(VIS)~\cite{yang2021crossvis,wu2022idol,huang2022minvis,ravi2024sam2} approaches used for this task provide temporal correspondence through motion, appearance, and spatial cues. However, accurate localization and short-term tracking do not necessarily yield accurate unique-instance counts. When a trajectory fragments, assigning a fresh identity to a previously counted object can inflate the count, while merging distinct objects under a single identity can suppress valid counts.

Open-vocabulary counting extends this task beyond predefined categories by allowing targets to be specified through natural-language prompts. In videos, this category flexibility must be paired with consistent identity assignment across frames~\cite{amini2025countvid,Owusu_2026_CVPR}. CountVid~\cite{amini2025countvid} introduced open-vocabulary video counting by combining CountGD-Box localization with SAM~2~\cite{ravi2024sam2} tracking and temporal consistency checks. However, relying on spatial overlap to identify new objects leaves CountVid vulnerable to identity fragmentation and repeated counting following occlusions or re-entry. More recently, SAM~3~\cite{carion2026sam3} unified concept-conditioned detection and multi-object tracking, allowing its output identities to serve as a direct counting baseline. Building on SAM~3, SAM3Count~\cite{Owusu_2026_CVPR} incorporates multimodal feature matching to recover lost identities during tracking. Despite these advances, both CountVid and SAM3Count remain sensitive to identity errors inherited from their underlying trackers.
Figure~\ref{fig:teaser} illustrates these limitations on a Dance-Count video featuring 7 labeled dancers. SAM~3 reports 8 identities within the first 5\% of the video. It continues issuing new identities as dancers overlap or re-enter the scene, ultimately reporting 14 identities. CountVid and SAM3Count struggle similarly, predicting 13 and 11 identities, whereas CIA reports 8.

To address this, we introduce Consistent Identity Association (\textbf{CIA}), a learned association module for open-vocabulary video object counting. Whenever the underlying tracker issues new identities, CIA jointly compares the corresponding detections with active and dormant identities using appearance, geometry, motion, and temporal cues. It distinguishes new objects, returning instances, and duplicates of active tracks, maintaining consistent counting identities across the video. For video counting evaluation, identity decisions require more than count metrics. Missed objects, duplicates, or false identities can offset each other. This produces an accurate total despite incorrect identity assignments. We therefore introduce the Video Counting Benchmark (VCB), comprising 959 videos and 1,647 queries with instance-level identity annotations. These annotations enable us to measure composition error that exposes failures concealed by the final count, and temporal identity metrics to assess whether predictions remain associated with the same objects across frames. Our contributions are:
\begin{itemize}
\item We formulate open-vocabulary video counting as an identity-assignment problem and decompose counting errors into missed, duplicate, and false identities.
\item We introduce CIA, a lightweight association module that jointly distinguishes new, returning, and duplicate tracks to maintain consistent identities.
\item We introduce VCB and evaluate CIA on VCB and VideoCount~\cite{amini2025countvid}, measuring count accuracy, composition error, and temporal identity consistency.
\end{itemize}

\section{Related Work}
\label{sec:related}
\paragraph{Video individual counting:} Video individual counting (VIC) estimates the number of distinct pedestrians observed throughout a video, requiring repeated observations of the same person to contribute only once. DR.VIC~\cite{Han_2022_CVPR} decomposes this count into the initial population and subsequent inflow, using cross-frame correspondence to identify newly appearing individuals. CGNet~\cite{Liu_2024_CVPR} reduces dependence on trajectory annotations through group-level contrastive learning from inflow and outflow supervision while maintaining individual templates to recognize reappearances. Subsequent methods incorporate self-supervised representation learning~\cite{huang2025vicssl}, social-group and spatiotemporal displacement priors for crowded scenes~\cite{lu2026omanpp}, and density-based correspondence under moving drone cameras~\cite{fan2025video,fan2026gd3a}. While these approaches establish temporal correspondence as central to distinct-instance counting, they focus exclusively on pedestrians. Our CIA addresses distinct-instance counting across text-specified categories.

\paragraph{Open-vocabulary video counting:} Open-vocabulary video counting extends distinct-instance counting to targets specified by text. CountVid~\cite{amini2025countvid} combines CountGD-Box detection with SAM~2~\cite{ravi2024sam2} mask propagation and temporal consistency filtering, counting instances through the resulting masklets. However, it lacks explicit re-identification after tracking continuity is lost. SAM~3~\cite{carion2026sam3} unifies concept-conditioned detection and multi-object tracking, allowing text-specified instances to be counted through its output identities. SAM3Count~\cite{Owusu_2026_CVPR} adds multimodal re-identification to recover lost identities. Nevertheless, fragmented trajectories, duplicate tracks, and incorrect identity assignments inherited from the underlying tracker can still affect the final count. Our CIA addresses these counting errors by jointly associating newly issued tracks with previously counted objects.

\begin{figure*}[!h]
\centering
\includegraphics[width=\textwidth]{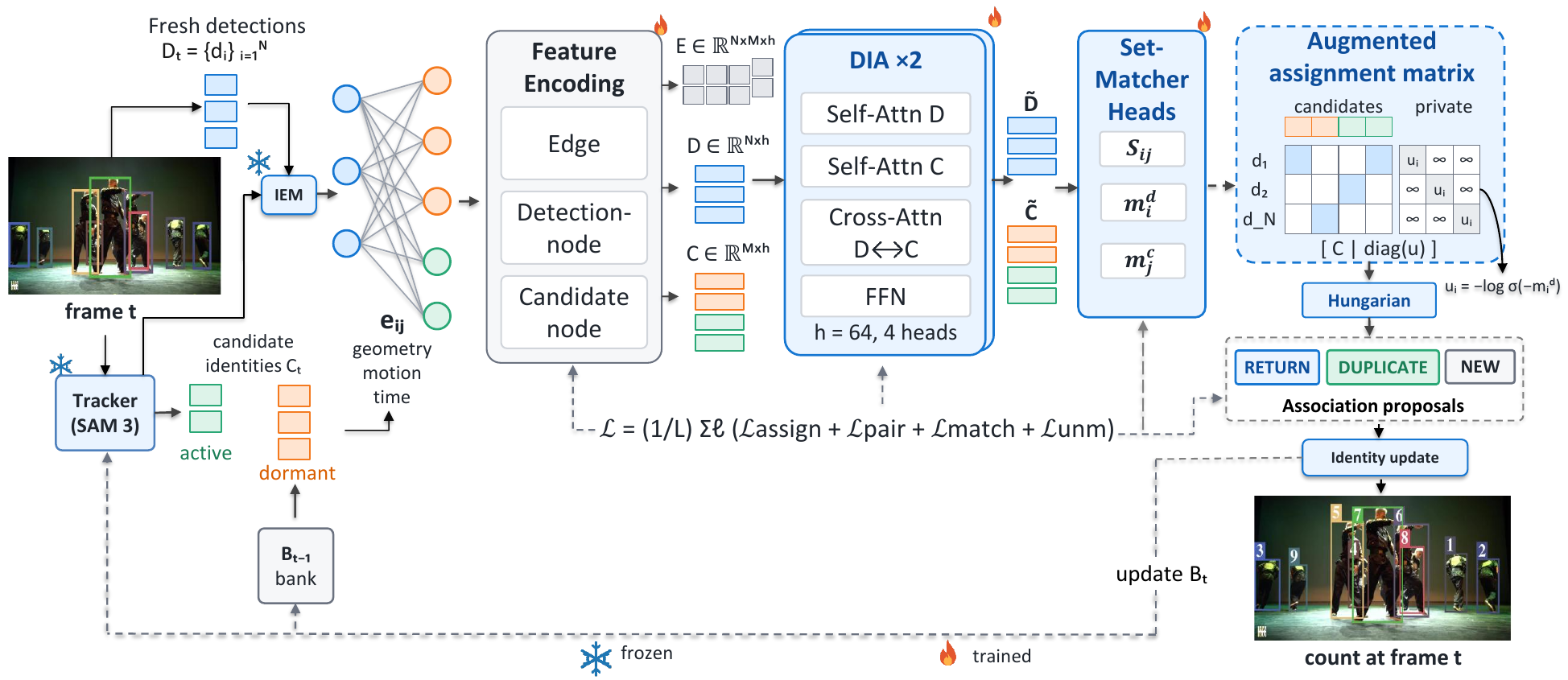}
\caption{CIA architecture. Fresh tracker detections are compared with active and dormant identities using IEM appearance descriptors, geometry, motion, and time cues. Feature encoding forms representations refined through DIA's attention blocks. Matching heads predict compatibility and matchability scores for augmented Hungarian assignment, and the resulting proposals are validated to update identities and the count.}
\label{fig:arch}
\end{figure*}

\paragraph{Identity association in tracking and segmentation:} MOT methods maintain temporal correspondence through motion and appearance cues~\cite{zhang2022bytetrack, Cao_2023_CVPR,aharon2022bot} or learned associations~\cite{Gao_2025_CVPR,adzemovic2026tdlp}. OVTrack~\cite{li2023ovtrack} and MASA~\cite{li2024masa} extend tracking to arbitrary categories, while AFLink~\cite{Du_2023_TMM} reconnects fragmented tracklets through appearance-free global association.
VIS methods similarly handle dynamic object populations. DVIS-DAQ~\cite{zhou2024dvisdaq} introduces dynamic anchor queries for emerging and disappearing instances, and LOMM~\cite{Lee_2025_LOMM} combines object memory with occupancy-guided matching to distinguish existing and newly appearing objects while keeping its tracker frozen. SAM2MOT~\cite{jiang2025sam2mot} extends SAM~2 to multi-object tracking through trajectory management and cross-object interactions. These methods address temporal association, track recovery, and identity continuity through matching and memory management. Our CIA focuses on determining whether newly issued tracker identities should increase the count. It jointly matches the corresponding detections against dormant and active identities, with an unmatched option for new objects. Matching a dormant identity supports re-entry, while matching an active identity identifies a potential duplicate, subject to validation.
\section{Method}
\label{sec:method}
\subsection{Problem Formulation and Overview}
\label{sec:method_overview}

Given a video $V=\{I_t\}_{t=1}^{T}$ and a text query $q$, our goal is to estimate the number $N_q$ of distinct objects matching $q$, with each physical object contributing exactly once. We formulate this as an identity-assignment problem to determine whether newly issued tracker identities correspond to already counted objects or new instances. CIA jointly compares the corresponding observations with identities that are currently tracked (active) or retained after tracking is interrupted (dormant). Unmatched observations can be classified as new objects or rejected observations.

We instantiate CIA using frozen SAM~3~\cite{carion2026sam3} as the underlying detector and tracker. CIA maintains consistent counting identities separately from SAM~3's native track IDs, updating them through joint assignment and subsequent identity validation. Let $\mathcal T_t$ denote the set of counting trajectories maintained at frame $t$, giving the current count $\widehat N_{q,t}=|\mathcal T_t|$. After processing the video, trajectory consolidation produces the final set $\mathcal T$ and predicted count $\widehat N_q=|\mathcal T|$. Figure~\ref{fig:arch} illustrates the overall architecture.

\subsection{Observations and Identity Representation}
\label{sec:background}
\label{sec:cia}
\paragraph{Observations and identity state.}
Frozen SAM~3 supplies query-conditioned masks, confidence scores, and native track IDs. CIA maps these native IDs to counting identities maintained in an identity bank $B_t$, which is initialized empty for each video--query pair. Each entry stores appearance evidence, recent boxes, and the last observation time. Continuing tracks retain their counting identities, while identities no longer tracked remain in the bank for possible recovery. CIA performs association when a fresh tracker ID appears or an existing track returns after the configured tracking gap.

At frame $t$, let $\mathcal D_t=\{d_i\}_{i=1}^{N}$ denote the observations requiring association. CIA compares them with candidate identities $\mathcal C_t=\{c_j\}_{j=1}^{M}$ comprising dormant identities retained in $B_{t-1}$ and active identities continued by the tracker in the current frame:
\begin{equation}
\mathcal C_t=
\mathcal C_t^{\mathrm{dormant}}
\cup
\mathcal C_t^{\mathrm{active}}.
\label{eq:fs_candidates}
\end{equation}
Dormant identities remain available for association within a fixed retention window. When they expire, they are removed from the candidate set without changing the accumulated count.

\paragraph{Appearance representation.}
We train an identity embedding module (IEM) to adapt SAM~3 features for distinguishing individual objects. Supervised contrastive learning brings observations of the same identity closer while separating different identities. The IEM projects each mask-pooled SAM~3 appearance feature $a_i\in\mathbb R^{256}$ into a normalized descriptor:
\begin{equation}
z_i=
\frac{f_{\mathrm{IEM}}(a_i)}
{\|f_{\mathrm{IEM}}(a_i)\|_2},
\qquad
z_i\in\mathbb R^{128}.
\label{eq:fs_iem}
\end{equation}
The same IEM encodes detections and candidate observations, providing appearance evidence for recognizing previously counted objects. It remains frozen during association training and inference. IEM training and feature extraction are detailed in the supplementary material.

\subsection{Feature Encoding and Contextual Association}
\label{sec:fs_graph}

\paragraph{Edge and node encoding.}
We encode both the relationship between each detection and candidate and the properties of each individual observation. For a detection--candidate pair $(i,j)$, the cue vector $v_{ij}\in\mathbb R^{16}$ captures appearance, geometry, motion, time, confidence, and competing matches. Each cue has a binary availability flag in $r_{ij}\in\{0,1\}^{16}$; unavailable cues and their flags are set to zero. The edge encoder combines these measurements and flags into an embedding of width $h=64$:
\begin{equation}
x_{ij}=[v_{ij};r_{ij}]\in\mathbb R^{32},
\qquad
e_{ij}=f_e(x_{ij})\in\mathbb R^h.
\label{eq:fs_features}
\end{equation}
For example, motion cues require previous observations, whereas geometric comparisons with active candidates use their current-frame boxes. The supplement provides the cue definitions and availability rules.

The node encoders then combine each observation's appearance and geometry with a summary of its pairwise relationships. This summary is obtained by mean- and max-pooling over its available edge embeddings. Detection nodes additionally receive the current box and confidence, while candidate nodes receive the last observed and predicted boxes. Availability flags mark missing attributes, and a learned active-state embedding distinguishes active candidates from dormant ones. The resulting edge features $\mathbf E\in\mathbb R^{N\times M\times h}$, detection tokens $\mathbf D\in\mathbb R^{N\times h}$, and candidate tokens $\mathbf C\in\mathbb R^{M\times h}$ form the inputs to DIA.

\paragraph{Dormant Identity Associator (DIA).}
The Dormant Identity Associator (DIA) jointly compares fresh detections with dormant and active identities. Following the contextual matching approach used for image keypoints~\cite{Lindenberger_2023_ICCV}, we associate tracker observations with counting identities across time. DIA uses self-attention to capture relationships within the detection and candidate sets, and cross-attention to compare detections with candidates. This contextual reasoning allows each potential match to be assessed against competing detections and alternative identities, producing refined representations $\widetilde{\mathbf D}$ and $\widetilde{\mathbf C}$ for the matching heads.

\subsection{Joint Assignment and Identity Updates}
\label{sec:fs_assignment}

\paragraph{Compatibility and assignment:} The matching heads predict pairwise compatibility scores $S_{ij}$ from the refined detection and candidate representations, along with edge evidence. Detection and candidate matchability logits, $m_i^d$ and $m_j^c$, estimate whether each observation or identity corresponds to an element in the opposing set. Following ~\cite{Lindenberger_2023_ICCV}, we combine dual-softmax scores with matchability, adding absolute pair evidence $\sigma(S_{ij})$:
\begin{align}
\log P_{ij}
&=
\log\operatorname{softmax}_{j}(S_{ij})
+\log\operatorname{softmax}_{i}(S_{ij})
\nonumber\\
&\quad+
\log\sigma(S_{ij})
+\log\sigma(m_i^d)
+\log\sigma(m_j^c).
\label{eq:score}
\end{align}
The absolute pair term provides evidence beyond relative ranking, including when only one candidate is available.

We minimize the real-candidate costs $-\log P_{ij}$ using Hungarian assignment, augmented with a private unmatched alternative for each detection at cost $u_i=-\log\sigma(-m_i^d)$. Each detection is assigned once and each candidate at most once. Unmatched alternatives allow observations to remain unassociated when no candidate is suitable.

\paragraph{Identity updates and counting.}
Assignment results propose how each observation should affect the count. A match to a dormant identity proposes a \emph{return}, reusing its count identity, while a match to an active identity proposes a \emph{duplicate}, subject to geometric validation. Unmatched observations are classified as new objects or rejected observations.
Before creating a new counting identity, proposals contained within an active track undergo short temporal confirmation to check for duplication. Duplicate suppression remains reversible. if a suppressed track later separates from its host, it can receive a new counting identity with its buffered observations restored.
These decisions update the identity bank $B_t$ and counting trajectories $\mathcal T_t$, giving the current count $\widehat N_{q,t}=|\mathcal T_t|$. After the video, appearance- and geometry-based consolidation merges remaining fragmented or duplicate trajectories to produce the final set $\mathcal T$ and count $\widehat N_q=|\mathcal T|$.

\subsection{Training Objective}
\label{sec:fs_training}

We train DIA on association events extracted from frozen tracker runs. Trajectory annotations label candidate correspondences, new objects, and rejected observations. DIA receives no category names or ground-truth identities as inputs.
The objective averages four equally weighted losses over the $L$ transformer blocks:
\begin{equation}
\mathcal L=\frac{1}{L}\sum_{\ell=1}^{L}
\bigl(
\mathcal L_{\mathrm{assign}}^\ell+
\mathcal L_{\mathrm{pair}}^\ell+
\mathcal L_{\mathrm{match}}^\ell+
\mathcal L_{\mathrm{unm}}^\ell
\bigr).
\label{eq:fs_loss}
\end{equation}
The assignment loss maximizes the scores of labeled correspondences. Pair validity uses balanced binary cross-entropy with hard-negative mining, while matchability supervises whether detections and candidates have valid correspondences. The unmatched loss distinguishes new objects from rejected observations without entering the Hungarian cost.

DIA is trained on OVIS~\cite{qi2022occluded} train split and DanceTrack~\cite{Sun_2022_CVPR} train split, with OVIS validation and held-out DanceTrack training videos used for development. The detector, tracker, and pretrained IEM remain frozen. Sampling, optimization, and model-selection details are provided in the supplement.

\section{Experiments}

\label{sec:experiments}
\begin{table*}[!h]
\centering\small\setlength{\tabcolsep}{3pt}
\caption{Counting results on VCB. CE: composition error; on BURST-Count, it is computed on BURST's labeled frames. Best in bold.}\label{tab:counting}
\resizebox{\textwidth}{!}{%
\begin{tabular}{lcccccccccccc}
\toprule
 & \multicolumn{3}{c}{BURST-Count} & \multicolumn{3}{c}{Dance-Count} & \multicolumn{3}{c}{Animal-Count} & \multicolumn{3}{c}{GMOT-Count} \\
\cmidrule(lr){2-4}\cmidrule(lr){5-7}\cmidrule(lr){8-10}\cmidrule(lr){11-13}
Method & MAE$\downarrow$ & RMSE$\downarrow$ & CE$\downarrow$ & MAE$\downarrow$ & RMSE$\downarrow$ & CE$\downarrow$ & MAE$\downarrow$ & RMSE$\downarrow$ & CE$\downarrow$ & MAE$\downarrow$ & RMSE$\downarrow$ & CE$\downarrow$ \\

\midrule
\multicolumn{13}{l}{\emph{Tracking-based controls}} \\
SAM~3~\cite{carion2026sam3} & 0.656 & 1.769 & 0.794 & 4.920 & 7.892 & 5.000 & -- & -- & -- & 11.421 & 25.252 & 14.947 \\
ByteTrack~\cite{zhang2022bytetrack} (ECCV'22) & 2.328 & 4.218 & 2.583 & 37.760 & 58.194 & 39.920 & 26.480 & 38.301 & 37.520 & 29.868 & 59.481 & 39.658 \\
OC-SORT~\cite{Cao_2023_CVPR} (CVPR'23) & 3.750 & 7.614 & 3.984 & 36.120 & 52.105 & 37.960 & 34.400 & 52.020 & 47.680 & 50.553 & 123.922 & 61.289 \\
BoT-SORT~\cite{aharon2022bot} (arXiv'22) & 2.197 & 4.044 & 2.445 & 32.360 & 49.587 & 33.960 & 23.960 & 34.492 & 34.760 & 26.947 & 53.279 & 36.263 \\
StrongSORT~\cite{Du_2023_TMM} (TMM'23) & 3.760 & 6.462 & 4.045 & 79.480 & 112.780 & 79.720 & 44.800 & 57.263 & 55.840 & 57.400 & 93.770 & 63.500 \\
Deep OC-SORT~\cite{maggiolino2023deep} (ICIP'23) & 5.756 & 10.228 & 5.967 & 83.800 & 116.952 & 83.880 & 67.640 & 92.429 & 77.400 & 83.475 & 159.970 & 88.875 \\
MASA~\cite{li2024masa} (CVPR'24) & 3.013 & 7.685 & 3.269 & 165.440 & 238.361 & 165.760 & 56.920 & 77.902 & 68.840 & 63.421 & 125.104 & 69.579 \\
\midrule
\multicolumn{13}{l}{\emph{Video counting systems}} \\
CountVid~\cite{amini2025countvid} (AAAI'26) & 1.421 & 4.680 & 1.756 & 3.800 & 5.378 & 4.760 & 8.760 & 14.611 & 18.440 & 12.211 & 26.199 & 20.158 \\
SAM3Count~\cite{Owusu_2026_CVPR} & 0.594 & 1.546 & 0.763 & 4.440 & 7.513 & 4.520 & -- & -- & -- & 13.263 & 26.896 & 18.684 \\
\textbf{CIA (ours)} & \textbf{0.575} & \textbf{1.403} & \textbf{0.725} & \textbf{1.400} & \textbf{2.324} & \textbf{2.600} & \textbf{7.20} & \textbf{10.68} & \textbf{15.68} & \textbf{10.684} & \textbf{24.407} & \textbf{14.842} \\
\bottomrule
\end{tabular}}
\end{table*}

We evaluate CIA on VCB and VideoCount~\cite{amini2025countvid}, comparing it with open-vocabulary video counting methods and MOT methods using their built-in association mechanisms on the same SAM~3 detections. We also test CIA as a plugin associator to alternative detectors and tracking pipelines without retraining. This examines its association and identity-consistency behavior beyond its baseline detector and tracker.

\begin{table}[!ht]\centering\small
\caption{Counting results on VideoCount~\cite{amini2025countvid}}
\label{tab:videocount}
\resizebox{\linewidth}{!}{%
\begin{tabular}{lcccc}
\toprule
 & \multicolumn{2}{c}{TAO-Count} & \multicolumn{2}{c}{Penguins} \\
\cmidrule(lr){2-3}\cmidrule(lr){4-5}
Method & MAE$\downarrow$ & RMSE$\downarrow$ & MAE$\downarrow$ & RMSE$\downarrow$ \\
\midrule
SAM~3~\cite{carion2026sam3} & 0.653 & 1.603 & 3.667 & 4.933 \\
CountVid~\cite{amini2025countvid} (AAAI'26) & 2.595 & 5.894 & 3.667 & 5.323 \\
SAM3Count~\cite{Owusu_2026_CVPR} & 0.78 & 1.63 & 2.667 & 3.162 \\
\midrule
\textbf{CIA (ours)} & \textbf{0.604} & \textbf{1.429} & \textbf{2.333} & \textbf{3.109 }\\
\bottomrule
\end{tabular}}
\end{table}
\begin{table}[!ht]
\centering\small
\caption{Counting on the Crystals dataset from VideoCount ~\cite{amini2025countvid}. SAM~3 here means its image detector and video tracker. "+CIA" is the association module integrated into the pipeline.}
\label{tab:crystals}
\resizebox{\linewidth}{!}{%
\begin{tabular}{lcccc}
\toprule
 & \multicolumn{2}{c}{Text} & \multicolumn{2}{c}{Text + internal exemplars} \\
\cmidrule(lr){2-3}\cmidrule(lr){4-5}
Method & MAE$\downarrow$ & RMSE$\downarrow$ & MAE$\downarrow$ & RMSE$\downarrow$ \\
\midrule
SAM~3 & 74.714 & 89.450 & 74.714 & 89.450 \\
SAM~3 + CIA & 74.714 & 89.450 & 74.714 & 89.450 \\
CountVid & 69.000 & 86.012 & 27.857 & 45.041 \\
CountVid + CIA  & 64.71 & 83.09 & 24.43 & 36.53\\
CountGD-box + SAM~3 + CIA & \textbf{64.714} & \textbf{83.086} & \textbf{19.571} & \textbf{32.896} \\
\bottomrule
\end{tabular}}
\end{table}

\begin{figure}[t]
\centering
\includegraphics[width=\columnwidth]{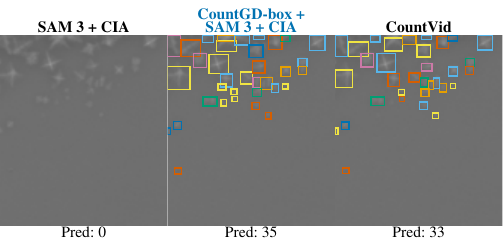}
\caption{Crystals frame with a ground-truth count of 37 and prompt: "white crystal". SAM~3+ CIA detects no targets, while CountGD-box + SAM~3 tracker with CIA and CountVid produce crystal observations. Pred denotes the number of output boxes in the frame.}
\label{fig:crystals-detector}
\end{figure}

\subsection{Datasets and metrics}
\label{sec:setup}
\paragraph{Datasets:} A query is one category in one video, and its count label is the number of distinct labeled track identities of that category. We evaluate on two benchmarks; their sizes, object counts, and frame rates are given in the supplement.

\noindent\textbf{VCB:} We introduce VCB, whose videos are derived from four existing datasets. We retain only video--category queries in which all target objects have annotated track identities. For GMOT-Count ~\cite{Bai_2021_CVPR}, we manually added names or short descriptions, such as ``person marching in red clothing,'' as text prompts for its videos. We then convert the annotations to per-frame boxes and evaluate counting, identity composition, and temporal correspondence using the same matching protocol. BURST-Count follows CountVid's TAO-Count filtering procedure~\cite{amini2025countvid} on BURST-test~\cite{Athar_2023_BURST}, yielding 1,556 queries across 868 videos and 274 categories. In Dance-Count, we use 25 DanceTrack validation videos~\cite{Sun_2022_CVPR}, prompted with ``dancer'' to exclude unlabeled spectators. Animal-Count uses 25 AnimalTrack test videos covering ten species~\cite{zhang2023animaltrack}, with species names as prompts. Details of per-dataset prompts and queries are in the supplementary material. We selected these datasets because they encompass a wide range of challenging scenarios, such as dense crowds, heavy occlusion, long tailed distributions, reentries, and lookalikes.

\noindent\textbf{VideoCount:} CountVid's benchmark~\cite{amini2025countvid} comprises TAO-Count, with 357 TAO~\cite{dave2020tao} validation videos and 639 queries; Penguins, with three videos of crowded penguin colonies; and Crystals, with seven microscopy videos of growing crystals. VideoCount provides only per-frame count annotations, so it supports only count evaluation. 
\begin{table*}[ht]
\centering
\footnotesize
\setlength{\tabcolsep}{2pt}
\caption{Identity metrics on VCB. Best per column in bold.}
\label{tab:identity}
\resizebox{\textwidth}{!}{%
\begin{tabular}{lcccccccccccccccc}
\toprule
 & \multicolumn{4}{c}{BURST-Count}
 & \multicolumn{4}{c}{Dance-Count}
 & \multicolumn{4}{c}{Animal-Count}
 & \multicolumn{4}{c}{GMOT-Count} \\
\cmidrule(lr){2-5}\cmidrule(lr){6-9}\cmidrule(lr){10-13}\cmidrule(lr){14-17}
Method
 & HOTA$\uparrow$ & AssA$\uparrow$ & IDF1$\uparrow$ & ID sw.$\downarrow$
 & HOTA$\uparrow$ & AssA$\uparrow$ & IDF1$\uparrow$ & ID sw.$\downarrow$
 & HOTA$\uparrow$ & AssA$\uparrow$ & IDF1$\uparrow$ & ID sw.$\downarrow$
 & HOTA$\uparrow$ & AssA$\uparrow$ & IDF1$\uparrow$ & ID sw.$\downarrow$ \\
\midrule
\multicolumn{17}{l}{\emph{Tracking-based controls}} \\
SAM~3~\cite{carion2026sam3}
 & 74.2 & \textbf{87.8} & 78.9 & 324
 & \textbf{72.6} & \textbf{72.5} & \textbf{80.7} & 1,715
 & -- & -- & -- & --
 & \textbf{64.3} & \textbf{72.7} & \textbf{79.0} & 1,075 \\
ByteTrack~\cite{zhang2022bytetrack} (ECCV'22)
 & 63.5 & 69.5 & 61.5 & 3,105
 & 53.7 & 45.2 & 53.2 & 2,348
 & 56.7 & \textbf{67.4} & 68.4 & 797
 & 57.4 & 66.2 & 67.1 & 2,018 \\
OC-SORT~\cite{Cao_2023_CVPR} (CVPR'23)
 & 62.9 & 67.2 & 59.8 & 4,503
 & 54.8 & 47.2 & 54.4 & 2,565
 & 54.7 & 63.0 & 65.2 & 1,045
 & 57.2 & 64.8 & 66.5 & 3,539 \\
BoT-SORT~\cite{aharon2022bot} (arXiv'22)
 & 63.9 & 70.4 & 62.2 & 2,939
 & 58.3 & 52.7 & 59.4 & 1,882
 & 56.7 & \textbf{67.4} & 68.7 & 736
 & 57.9 & 67.1 & 68.0 & 1,786 \\
StrongSORT~\cite{Du_2023_TMM} (TMM'23)
 & 58.8 & 62.5 & 57.8 & 3,416
 & 50.2 & 34.4 & 46.4 & 6,053
 & 55.4 & 56.3 & 64.8 & 1,971
 & 58.7 & 63.8 & 70.0 & 3,771 \\
Deep OC-SORT~\cite{maggiolino2023deep} (ICIP'23)
 & 60.2 & 60.5 & 57.0 & 5,967
 & 38.5 & 21.2 & 32.1 & 8,245
 & 51.8 & 49.6 & 59.3 & 2,987
 & 57.1 & 59.3 & 66.4 & 6,308 \\
MASA~\cite{li2024masa} (CVPR'24)
 & 64.1 & 72.8 & 65.5 & 2,750
 & 44.9 & 28.4 & 39.3 & 5,707
 & 56.5 & 58.9 & 66.9 & 1,499
 & 59.8 & 63.8 & 71.2 & 4,256 \\
\midrule
\multicolumn{17}{l}{\emph{Video counting systems}} \\
CountVid~\cite{amini2025countvid} (AAAI'26)
 & 48.1 & 78.3 & 46.0 & 583
 & 62.0 & 65.1 & 63.6 & 13,045
 & 50.5 & 58.9 & 57.2 & 8,348
 & 59.9 & 69.2 & 69.9 & 6,617 \\
SAM3Count~\cite{Owusu_2026_CVPR}
 & 71.5 & 84.9 & 77.0 & 310
 & 69.4 & 71.1 & 77.1 & 2,469
 & -- & -- & -- & --
 & 59.1 & 71.3 & 71.3 & 1,513 \\
\textbf{CIA (ours)}
 & \textbf{74.4} & 87.6 & \textbf{79.2} & \textbf{266}
 & 70.4 & 72.4 & 78.9 & \textbf{1,032}
 & \textbf{57.4} & 64.4 & \textbf{70.0} & \textbf{483}
 & 63.5 & 72.5 & 77.8 & \textbf{846} \\
\bottomrule
\end{tabular}}
\end{table*}
\begin{figure*}[!t]
\centering
\includegraphics[width=\textwidth]{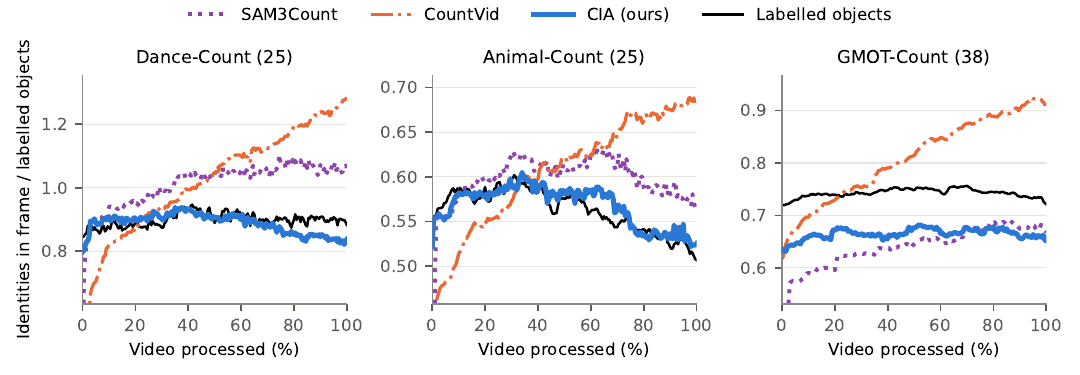}
\caption{Number of identities per frame, normalized by the query’s total labeled object count and averaged across queries, as video progress. The black line shows the corresponding ground-truth fraction.}
\label{fig:tracks_per_frame}
\end{figure*}
\paragraph{Metrics}
We report count MAE and RMSE over video--query pairs. An accurate total can conceal incorrect identities because extra trajectories can offset missed objects. For each video--query pair, we assign each predicted trajectory to the labeled object it overlaps most frequently at a box IoU of at least $0.5$. Trajectories with no qualifying overlap remain unassigned. Let $F$ denote unassigned trajectories, $D$ additional trajectories assigned to an already represented object, and $M_{\mathrm{miss}}$ labeled objects with no assigned trajectory. The signed count error is
$\widehat N_q-N_q=F+D-M_{\mathrm{miss}}.
\label{eq:decomp}$
To expose errors that cancel in the final count, we define composition error over $Q$ video--query pairs as
\begin{equation}
\mathrm{CE}=\frac{1}{Q}\sum_{r=1}^{Q}
\left(F_r+D_r+M_{\mathrm{miss},r}\right),
\end{equation}
where $r$ indexes the video--query pairs.
We report HOTA, AssA~\cite{luiten2021hota}, IDF1~\cite{Ristani_2016_IDF1}, and identity switches to assess temporal correspondence. 

\begin{figure*}[!ht]
\centering
\includegraphics[width=\textwidth]{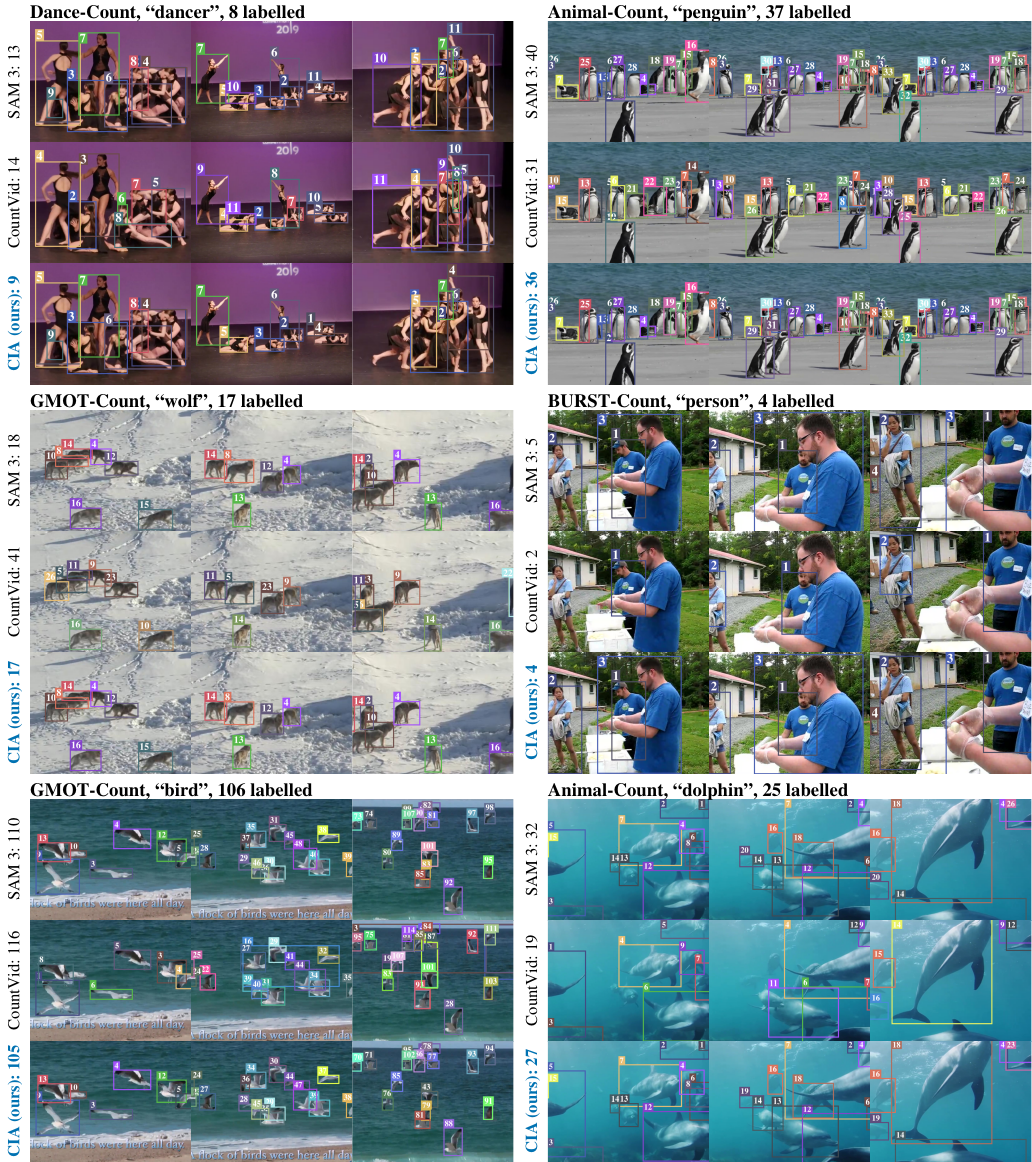}
\caption{Qualitative comparison on selected videos from the VCB sets, covering different scenarios. The prompt and number of labeled objects appear above each example, and each row shows a method and the number of identities it reports over the video.}
\label{fig:qual}
\end{figure*}
\subsection{Quantitative results}
\label{sec:results}
\paragraph{Counting:}
CIA achieves the lowest count metrics score on all VCB sets (Table~\ref{tab:counting}), with a particularly strong advantage over trackers using the same detections on Dance-Count. 
Among the compared systems, CIA has the lowest composition error on BURST-Count, Dance-Count, and GMOT-Count. The gap between CE and MAE, especially on Animal-Count, shows that extra identities can offset missed objects. An accurate total, therefore, does not necessarily mean that objects have been counted correctly.
\paragraph{Identity consistency:}\label{sec:identity}
CIA achieves the highest HOTA and IDF1 on BURST-Count and Animal-Count, and the fewest identity switches on every VCB set (Table~\ref{tab:identity}). On Dance-Count and GMOT-Count, CIA ranks second to SAM~3 in HOTA and IDF1, while reducing SAM~3's identity switches by approximately 40\% and 21\%, respectively. These results show that CIA consistently improves identity continuity, even when SAM~3 retains an advantage in aggregate tracking quality. On Animal-Count, ByteTrack and BoT-SORT obtain higher AssA, but CIA achieves higher HOTA and IDF1 with substantially fewer identity switches. Because HOTA and IDF1 reflect both observation coverage and association quality, reducing identity switches does not necessarily improve every correspondence metric when objects are missed or inaccurately localized.

\paragraph{Scalability:} Under our evaluation configuration, the released SAM~3 and SAM3Count implementations exceed the memory capacity of an 80\,GB H100 on dense videos in MOT20-Count (52--398 processed frames and 80--1,203 ground-truth objects per video) and AnimalTrack-Count (196--2,126 processed frames and 8--134 ground-truth objects per video). We therefore do not report their results on these benchmarks. Figure~\ref{fig:tracks_per_frame} compares the number of predicted identities present in each frame with the labeled object population as the video progresses. On Dance-Count, CIA remains close to the labeled-object curve, whereas CountVid increasingly exceeds it as duplicate identities accumulate. On Animal-Count, CIA follows the labeled population more closely, while CountVid transitions from incomplete coverage to a slight excess. On GMOT-Count, CIA remains below the labeled curve, indicating that detector and tracker coverage remains a limiting factor even when temporal identity consistency improves.

\paragraph{Detector dependence:} Table~\ref{tab:crystals} and Figure~\ref{fig:crystals-detector} illustrate CIA's dependence on detector coverage. The SAM~3 detector produces no observations in either the text-only or internal-exemplar setting; consequently, SAM~3 + CIA report no objects in these videos. Because CIA operates on tracker observations, it cannot recover objects that the detector misses. Replacing the detector with CountGD-box~\cite{amini2025countvid} supplies crystal observations and reduces counting error while retaining the SAM~3 tracker and CIA association model. We also integrate CIA with CountVid's detection-and-tracking pipeline, and it performs better than the baseline. This shows that CIA can associate observations produced by a pipeline it was not trained on. In the internal-exemplar setting, high-confidence detector predictions are automatically selected as exemplar boxes and combined with the text prompt. These experiments demonstrate CIA's portability across detection and tracking pipelines.

\begin{table}[!h]

\centering\footnotesize
\caption{Ablations of CIA's principal components on Animal-Count. Rows marked $^{r}$ retrain the association network.}
\label{tab:ablation}
\resizebox{0.96\columnwidth}{!}{%
\begin{tabular}{lccc}
\toprule
Variant & MAE$\downarrow$ & CE$\downarrow$ & IDF1$\uparrow$ \\
\midrule
CIA & \textbf{7.640} & \textbf{15.640} & \textbf{69.8} \\
Without IEM$^{r}$ & 8.680 & 16.520 & 69.4 \\
\midrule
Without $\mathcal{L}_{\mathrm{match}}$$^{r}$ & 15.480 & 19.240 & 65.1 \\
Without $\mathcal{L}_{\mathrm{unm}}$$^{r}$ & 40.440 & 40.440 & 0.0 \\
Without $\mathcal{L}_{\mathrm{pair}}$$^{r}$
& 7.720 & 15.880 & 69.7 \\
\bottomrule
\end{tabular}}
\end{table}

\paragraph{Cost:} The IEM is pretrained separately and frozen while DIA is trained. Excluding the underlying detector and tracker, CIA contains 333,702 learned parameters. 99,584 in the IEM and 234,118 in DIA. On an NVIDIA H100, the DIA forward pass and augmented Hungarian assignment require 0.96\,ms per association-event frame.

\subsection{Ablation study}
\label{sec:ablations}
Table~\ref{tab:ablation} evaluates the IEM and the two losses most consequential to association. IEM improves count accuracy and composition error, while removing matchability supervision substantially degrades all three metrics. Without unmatched-observation supervision, the model fails to distinguish new objects from rejected observations. 

\subsection{Qualitative results}
\label{sec:qualitative}
Figure~\ref{fig:qual} shows how CIA addresses different identity-association failures. In Dance-Count, frequent overlap and changing configurations fragment the dancers' trajectories: SAM~3 and CountVid report 13 and 14 identities for 8 dancers, whereas CIA reports 9. On the GMOT wolf video, CountVid produces 41 identities for 17 animals, while CIA preserves identities across the selected views and recovers the correct count of 17. CIA also recovers the correct count of 4 in the BURST person example, where SAM~3 overcounts by one, and CountVid misses two identities.

The remaining examples test association in dense groups of visually similar objects. CIA reports 36 identities for 37 penguins and 105 for 106 birds, reducing the excess identities produced by SAM~3 in both cases and the larger overcount from CountVid on the bird video. In the dolphin example, CIA reduces SAM~3's count from 32 to 27 for 25 labeled animals, but does not remove every duplicate; CountVid instead undercounts with 19. These cases illustrate the central trade-off: CIA must reconnect fragmented or duplicate trajectories without merging distinct nearby objects. Errors remain when appearance and motion are ambiguous, and association cannot recover objects absent from the detector's observations.

\section{Conclusion}
\label{sec:conclusion}
We presented CIA, a lightweight association module for consistent identity assignment in open-vocabulary video counting. CIA compares fresh detections with dormant and active identities jointly to distinguish new objects from returning and duplicate observations.   We also introduced VCD, an evaluation protocol, and composition error to expose missed objects and redundant identities that affect the final count. Experiments show that CIA reduces repeated counting and identity switches across occlusions and re-entries. It reduces counting errors over CountVid and SAM3Count across the evaluated VCD and VideoCount sets.  Reliable counting nevertheless remains dependent on detection coverage. As shown in Fig~\ref{fig:crystals-detector}, CIA cannot recover objects that the frozen detector misses throughout the video. Incorrect detector localization can also lead to incorrect associations.

{
    \small
    \bibliographystyle{ieeenat_fullname}
    \bibliography{main}
}


\clearpage
\setcounter{page}{1}
\maketitlesupplementary
\appendix
\setcounter{table}{0}
\renewcommand{\thetable}{S\arabic{table}}

\section{Implementation Details}
\label{sec:supp-architecture}

\subsection{Identity Embedding and Feature Encoding}
\paragraph{Identity embedding:} The identity embedding module (IEM) takes a 256-dimensional SAM~3 feature averaged within the predicted object mask. Its appearance branch first projects this feature using layer normalization, a $256\!\rightarrow\!128$ linear layer, and GELU. The resulting representation then passes through layer normalization, a $128\!\rightarrow\!256$ linear layer, GELU, dropout (rate 0.1), and a final $256\!\rightarrow\!128$ linear projection, followed by $\ell_2$ normalization. This normalized descriptor provides appearance representations for both pairwise cues and node inputs.

\paragraph{Pairwise cues:} Table~\ref{tab:supp_features} lists the 16 measurements computed for each detection--candidate pair. We append one availability bit to each measurement, producing a 32-dimensional edge input. We set an unavailable measurement to zero, while its availability bit records that it is missing. Box coordinates are normalized by image width and height; center distance and motion residuals are then normalized by the diagonal of this coordinate system. Finally, we divide the time gap and track age by 30\,s and 300\,s, respectively, and clip them to 1.

\begin{table}[!h]
\centering\small
\setlength{\tabcolsep}{3pt}
\caption{Edge measurements. Each cue has an availability flag.}
\label{tab:supp_features}
\begin{tabular}{@{}lp{0.68\columnwidth}@{}}
\toprule
Group & Measurements \\
\midrule
Appearance & Cosine similarity; margin over the best alternative candidate \\
Geometry & Mask IoU; box IoU; center distance; absolute log area ratio; absolute log aspect-ratio difference \\
Motion/time & Motion residual; time gap; track age \\
Reliability & Detection confidence; tracker confidence; recent occlusion \\
Competition & Column margin; reciprocal best match; competitor count \\
\bottomrule
\end{tabular}
\end{table}

Appearance margins measure how a pair's cosine similarity compares with the strongest alternative candidate or detection. Reciprocal matching marks mutual best appearance matches, and competitor count records the number of feasible detections for a candidate. Motion uses up to five recent boxes and is unavailable beyond a 10\,s extrapolation horizon. For active candidates, current-frame boxes are used; mask-overlap and velocity-residual cues are unavailable.

\subsection{Contextual Association Network}
Table~\ref{tab:supp_architecture} gives DIA's layer dimensions. Separate projections map IEM descriptors to a shared 64-dimensional token space, while each node combines its attributes with mean- and max-pooled incident edges; empty pools are zero. A learned state embedding identifies active candidate tokens.

\begin{table}[!h]
\centering\small
\setlength{\tabcolsep}{3pt}
\caption{DIA layer dimensions. Arrows denote linear maps, with GELU between successive maps. Normalization is specified in the text. Blocks have separate weights; prediction heads are shared.}
\label{tab:supp_architecture}
\begin{tabular}{@{}ll@{}}
\toprule
Module & Dimensions \\
\midrule
Edge encoder & $32\!\rightarrow\!64\!\rightarrow\!64$ \\
Appearance projection (per set) & $128\!\rightarrow\!64$ \\
Detection node encoder & $199\!\rightarrow\!64$ \\
Candidate node encoder & $203\!\rightarrow\!64$ \\
Attention projections & $64\!\rightarrow\!64$ \\
Feed-forward network (per set) & $64\!\rightarrow\!128\!\rightarrow\!64$ \\
Pair-score projections (per set) & $64\!\rightarrow\!64$ \\
Edge-bias head & $64\!\rightarrow\!32\!\rightarrow\!1$ \\
Detection matchability & $71\!\rightarrow\!32\!\rightarrow\!1$ \\
Candidate matchability & $69\!\rightarrow\!32\!\rightarrow\!1$ \\
Auxiliary unmatched heads (each) & $65\!\rightarrow\!32\!\rightarrow\!1$ \\
\bottomrule
\end{tabular}
\end{table}

DIA uses two four-head transformer blocks. Each block performs self-attention within the detection and candidate sets, then cross-attention from candidates to detections and from the updated detections back to candidates. Residual, post-normalization, and feed-forward updates follow each attention operation; masks exclude padding and infeasible pairs. Edge embeddings remain fixed, and prediction heads are shared across blocks.

The matchability heads use refined node tokens and summary statistics of their incident pair scores. The unmatched classifier uses the detection token and confidence to distinguish new objects from rejected observations after assignment; its logits are separate from the private-unmatched cost used for assignment.

\subsection{Training}
\label{sec:supp-training}
\paragraph{IEM pretraining:} We pretrain the IEM using supervised contrastive learning on BURST training features, with BURST validation features used for validation. The contrastive temperature is 0.05, the learning rate is $3\times10^{-4}$, and the weight decay is 0.01. Batches contain 16 identities with eight observations each. Training runs for 15 epochs with 500 steps per epoch. The IEM remains frozen during DIA training and inference.

\paragraph{Association training:} DIA is trained on association events from OVIS and DanceTrack training videos. We use stratified event frame sampling and AdamW with a learning rate $10^{-3}$, weight decay $10^{-4}$, and eight event frames per batch. Training runs for at most 100 epochs, with gradient clipping at norm 1, node-appearance dropout 0.3, and early stopping after 15 epochs without improvement. Pair-validity supervision uses up to five highest-scoring negative candidates per detection. All four loss terms receive equal weight and are applied after both transformer blocks.


\subsection{Identity Memory and Trajectory Updates}
\label{sec:supp-method}
\paragraph{Candidate memory.}
At a processing rate of $f$ frames per second, an identity enters the dormant candidate set after $\max(1,\operatorname{round}(0.167f))$ missing frames and remains eligible for 30\,s. Aliases are consolidated before assignment. Once expired, an entry cannot be re-associated, though its historical contribution to the count remains intact.

\paragraph{Duplicate confirmation.}
An active match is accepted as a duplicate when the box IoU is at least 0.3, or the containment is at least 0.7, where containment measures the fraction of the detection box enclosed by the host box. Any proposed new detection satisfying this containment criterion is buffered for five subsequent observations. Each observation supports duplication if containment remains above the threshold and relative motion stays within 0.25 times the host box diagonal. If fewer than half of the buffered observations support duplication, we create a new identity and restore the buffered history.

Duplicate suppression remains reversible. A suppressed duplicate separates when both containment and IoU fall below 0.3, or when its relative motion exceeds 0.25 times the host box diagonal. Once separation persists for $\max(1,\operatorname{round}(f))$ processed observations, we create a new identity and restore its buffered history.

\paragraph{Trajectory consolidation.}
For temporally disjoint trajectories $a$ and $b$, identity-level consolidation requires a gap $\Delta t_{ab}\leq3$\,s, appearance similarity $s_{ab}\geq0.8$, endpoint-center distance $d_{ab}\leq\max(h_a,h_b)$, and height ratio $\max(h_a/h_b,h_b/h_a)\leq2$. Each trajectory has at most one predecessor and successor. Final geometric consolidation merges overlapping trajectories when they share at least five frames and their mean mask IoU exceeds $0.5$, computed on at most 20 shared frames. For short handovers with fewer than three shared frames, the endpoint gap must satisfy $\Delta f_{ab}\leq5$ processed frames, and the mask-center distance must be below $50$ native-resolution pixels. When trajectories overlap, the earlier trajectory retains the shared observations.

\section{Datasets and Evaluation Protocol}
\label{sec:supp-data}

\subsection{Dataset Coverage and Queries}
Table~\ref{tab:bench} summarizes the evaluation sets. Animal-Count uses 25 AnimalTrack test videos, excluding \texttt{duck\_3}. GMOT-Count contains 40 videos. Dance-Count uses the native 20\,fps rate, while Animal-Count and GMOT-Count use 30\,fps. OVIS-test is an additional association diagnostic, separate from VCB.

\begin{table*}[t]
\centering\small
\setlength{\tabcolsep}{4pt}
\caption{Evaluation coverage and processing rates. GMOT-Count contains 40 videos; matched comparisons use 38 after setting aside \texttt{balloon-3} and \texttt{car-1}. Crystals uses all supplied frames at about 6\,fps.}
\label{tab:bench}
\begin{tabular}{llrrrl}
\toprule
Dataset & Source & Videos & Queries & FPS & Query \\
\midrule
BURST-Count & BURST test~\cite{Athar_2023_BURST} & 868 & 1,556 & 3 & Exhaustively labeled category \\
Dance-Count & DanceTrack validation~\cite{Sun_2022_CVPR} & 25 & 25 & 20 & Dancer \\
Animal-Count & AnimalTrack test~\cite{zhang2023animaltrack} & 25 & 25 & 30 & Species name \\
GMOT-Count & GMOT-40~\cite{Bai_2021_CVPR} & 40 & 40 & 30 & Target name or description \\
TAO-Count & TAO validation~\cite{dave2020tao} & 357 & 639 & 3 & Target category \\
Penguins & VideoCount~\cite{amini2025countvid} & 3 & 3 & 3 & Penguin \\
Crystals & VideoCount~\cite{amini2025countvid} & 7 & 7 & $\approx6$ & White crystal \\
\bottomrule
\end{tabular}
\end{table*}

\begin{table*}[!h]
\centering\small
\setlength{\tabcolsep}{3pt}
\caption{Component ablations on VCB. $r$ marks variants that retrain DIA. $\dagger$: no counted identities are produced, so zero identity switches indicate collapse.}
\label{tab:supp_components_extended}
\resizebox{\textwidth}{!}{%
\begin{tabular}{lrrrrrrrrrrrr}
\toprule
 & \multicolumn{4}{c}{Dance-Count} & \multicolumn{4}{c}{Animal-Count} & \multicolumn{4}{c}{GMOT-Count} \\
\cmidrule(lr){2-5}\cmidrule(lr){6-9}\cmidrule(lr){10-13}
Variant & MAE$\downarrow$ & CE$\downarrow$ & IDF1$\uparrow$ & ID sw.$\downarrow$ & MAE$\downarrow$ & CE$\downarrow$ & IDF1$\uparrow$ & ID sw.$\downarrow$ & MAE$\downarrow$ & CE$\downarrow$ & IDF1$\uparrow$ & ID sw.$\downarrow$ \\
\midrule
CIA & 1.400 & 2.600 & 78.9 & 1,032 & 7.640 & 15.640 & 69.8 & 483 & 10.684 & 14.842 & 77.8 & 846 \\
Without active candidates$^{r}$ & 1.520 & 2.800 & 79.2 & 1,081 & 9.440 & 16.320 & 69.7 & 465 & 10.868 & 15.079 & 63.3 & 855 \\
Without IEM$^{r}$ & 1.880 & 3.400 & 78.4 & 1,068 & 8.680 & 16.520 & 69.4 & 484 & 11.895 & 15.579 & 77.0 & 766 \\
Without $\mathcal L_{\mathrm{assign}}$$^{r}$ & 2.240 & 3.520 & 78.3 & 1,152 & 8.720 & 16.160 & 70.5 & 501 & 10.684 & 14.526 & 77.7 & 814 \\
Without $\mathcal L_{\mathrm{pair}}$$^{r}$ & 1.800 & 3.160 & 79.6 & 1,078 & 7.720 & 15.880 & 69.6 & 519 & 10.553 & 14.184 & 78.2 & 814 \\
Without $\mathcal L_{\mathrm{match}}$$^{r}$ & 1.320 & 2.840 & 78.8 & 1,047 & 15.480 & 19.240 & 65.1 & 467 & 16.816 & 19.553 & 73.4 & 695 \\
Without $\mathcal L_{\mathrm{unm}}$$^{r,\dagger}$ & 10.920 & 10.920 & 0.0 & 0 & 40.440 & 40.440 & 0.0 & 0 & 47.921 & 47.921 & 0.0 & 0 \\
Negative budget 1,000$^{r}$ & 2.040 & 3.080 & 79.9 & 1,007 & 7.520 & 15.840 & 70.8 & 526 & 10.237 & 14.026 & 78.5 & 817 \\
\bottomrule
\end{tabular}}
\end{table*}

Each query specifies the exhaustively annotated target population in one video. BURST-Count retains only video--category pairs with exhaustive annotations; Dance-Count uses ``dancer'' to exclude unlabeled spectators; and Animal-Count uses the species name. GMOT prompts name the labeled population and add attributes when needed to distinguish it from other objects.

\subsection{Counting and Identity Evaluation}
\label{sec:supp-protocol}
We compute count error over all video--query pairs. To evaluate identity composition, we assign each predicted trajectory to the labeled object with the most box matches at IoU $\geq0.5$. Trajectories with no qualifying match remain unassigned. Composition error counts unassigned trajectories, duplicate trajectories assigned to an already represented object, and labeled objects with no assigned trajectory. Thus, zero composition error establishes a one-to-one trajectory assignment under this rule, but not temporal identity consistency.

Where trajectory annotations are available, HOTA and AssA are averaged over box-IoU thresholds from $0.05$ to $0.95$ in steps of $0.05$. IDF1 uses a global trajectory assignment and pools identity true positives, false positives, and false negatives across queries. We count an identity switch whenever the predicted identity matched to a labeled object changes, then sum switches over queries. HOTA, AssA, and IDF1 are reported as percentages throughout the supplement.

\subsection{Baseline Implementations}
\label{sec:supp-impl}
ByteTrack, OC-SORT, and motion-only BoT-SORT use BoxMOT 25.0.0 with a 30\,s buffer, a detection threshold of $0.7$, and low-score association disabled. Their settings are fixed on the DanceTrack development and AnimalTrack training videos. ByteTrack and BoT-SORT use an IoU-distance threshold of $0.9$; OC-SORT uses an IoU threshold of $0.2$, one minimum hit, $\Delta t=3$, and inertia $0.2$. BoT-SORT uses a proximity threshold of $0.5$ without camera-motion compensation.

CountVid uses its released pipeline at the stated processing rates. MASA uses its released association model on shared SAM~3 detections with a 30\,s memory. 

\section{Additional Ablation Studies}
\label{sec:supp-ablations}

\subsection{Appearance and Contextual Association}
Table~\ref{tab:supp_components_extended} extends the main-paper ablations to Dance-Count, Animal-Count, and GMOT-Count. The appearance ablation replaces IEM descriptors with raw SAM~3 features and retrains DIA. The active-candidate ablation retrains DIA without active identity columns. Both increase count error on Dance-Count and Animal-Count, supporting the use of adapted appearance and the full candidate state.

\subsection{Training Objective} The loss ablations in Table~\ref{tab:supp_components_extended} remove each objective term independently. Matchability supervision is particularly important for count accuracy on Animal-Count and GMOT-Count. Removing assignment supervision increases count error on Dance-Count and Animal-Count but slightly reduces it on GMOT-Count. Without unmatched-observation supervision, the model produces no counted identities; thus, its zero switches accompany zero HOTA and IDF1. This failure shows why switches must be assessed alongside coverage and counting metrics.

The negative-budget experiment increases the maximum number of highest-scoring negatives per detection from five to 1,000. It improves Animal-Count and GMOT-Count MAE but worsens Dance-Count MAE. It therefore measures the effect of a larger negative budget rather than the removal of negative supervision.

\subsection{Memory and Duplicate Confirmation}
Table~\ref{tab:supp_inference_sensitivity} varies inference settings while keeping the trained CIA network fixed. Changing retention also changes the time-gap normalization, so these rows evaluate the implemented memory configuration. The tested retention settings produce similar outcomes. Extending confirmation to 15 observations improves Dance-Count MAE but slightly worsens Animal-Count MAE. Disabling confirmation increases Animal-Count and GMOT-Count switches, although Dance-Count switches decrease slightly. CIA uses 30\,s retention and five-observation confirmation throughout the remaining experiments.

\begin{table*}[!h]
\centering\small
\setlength{\tabcolsep}{3pt}
\caption{Inference sensitivity with a fixed checkpoint on the same sets as Table~\ref{tab:supp_components_extended}. Retention changes also rescale the time-gap cue.}
\label{tab:supp_inference_sensitivity}
\resizebox{\textwidth}{!}{%
\begin{tabular}{lrrrrrrrrr}
\toprule
 & \multicolumn{3}{c}{Dance-Count} & \multicolumn{3}{c}{Animal-Count} & \multicolumn{3}{c}{GMOT-Count} \\
\cmidrule(lr){2-4}\cmidrule(lr){5-7}\cmidrule(lr){8-10}
Variant & MAE$\downarrow$ & CE$\downarrow$ & ID sw.$\downarrow$ & MAE$\downarrow$ & CE$\downarrow$ & ID sw.$\downarrow$ & MAE$\downarrow$ & CE$\downarrow$ & ID sw.$\downarrow$ \\
\midrule
CIA (30 s; confirmation 5) & 1.400 & 2.600 & 1,032 & 7.640 & 15.640 & 483 & 10.684 & 14.842 & 846 \\
Retention 10 s & 1.440 & 2.560 & 1,024 & 7.520 & 15.680 & 476 & 10.605 & 14.711 & 847 \\
Retention 40 s & 1.400 & 2.600 & 1,027 & 7.640 & 15.640 & 483 & 10.684 & 14.842 & 846 \\
Retention 50 s & 1.400 & 2.600 & 1,023 & 7.640 & 15.640 & 483 & 10.684 & 14.842 & 846 \\
Confirmation 15 observations & 1.280 & 2.640 & 1,023 & 7.680 & 15.520 & 490 & 10.684 & 14.789 & 834 \\
Without confirmation & 1.440 & 2.880 & 1,028 & 7.640 & 15.720 & 527 & 10.737 & 14.947 & 863 \\
Without dormant candidates & 3.840 & 4.080 & 1,150 & 7.600 & 15.200 & 515 & 10.263 & 14.263 & 914 \\
Without final merge & 1.480 & 2.680 & 1,057 & 7.600 & 15.760 & 483 & 10.684 & 14.895 & 849 \\
\bottomrule
\end{tabular}}
\end{table*}

\begin{table}[!h]
\centering\small
\caption{Final trajectory consolidation on BURST-Count and TAO-Count. Each pair shows CIA / CIA without final consolidation.}
\label{tab:supp_merge_additional}
\begin{tabular}{lcc}
\toprule
Dataset & MAE$\downarrow$ & RMSE$\downarrow$ \\
\midrule
BURST-Count & 0.575 / 0.579 & 1.403 / 1.412 \\
TAO-Count & 0.604 / 0.606 & 1.429 / 1.453 \\
\bottomrule
\end{tabular}
\end{table}

\begin{table*}[!h]
\centering\small
\setlength{\tabcolsep}{2.5pt}
\caption{Tracker portability. Each tracker is shown alone and with CIA applied to its outputs, without retraining CIA.}
\label{tab:supp_tracker_transfer}
\resizebox{\textwidth}{!}{%
\begin{tabular}{lrrrrrrrrrrrrrrr}
\toprule
 & \multicolumn{5}{c}{Dance-Count} & \multicolumn{5}{c}{Animal-Count} & \multicolumn{5}{c}{GMOT-Count} \\
\cmidrule(lr){2-6}\cmidrule(lr){7-11}\cmidrule(lr){12-16}
Tracker & MAE$\downarrow$ & CE$\downarrow$ & HOTA$\uparrow$ & IDF1$\uparrow$ & ID sw.$\downarrow$ & MAE$\downarrow$ & CE$\downarrow$ & HOTA$\uparrow$ & IDF1$\uparrow$ & ID sw.$\downarrow$ & MAE$\downarrow$ & CE$\downarrow$ & HOTA$\uparrow$ & IDF1$\uparrow$ & ID sw.$\downarrow$ \\
\midrule
ByteTrack & 37.760 & 39.920 & 53.7 & 53.2 & 2,348 & 26.480 & 37.520 & 56.7 & 68.4 & 797 & 29.868 & 39.658 & 57.4 & 67.1 & 2,018 \\
ByteTrack + CIA & 3.720 & 7.560 & 48.1 & 49.7 & 954 & 9.080 & 22.840 & 53.0 & 62.6 & 388 & 19.053 & 26.895 & 55.9 & 65.9 & 1,487 \\
\midrule
OC-SORT & 36.120 & 37.960 & 54.8 & 54.4 & 2,565 & 34.400 & 47.680 & 54.7 & 65.2 & 1,045 & 50.553 & 61.289 & 57.2 & 66.5 & 3,539 \\
OC-SORT + CIA & 5.080 & 8.440 & 50.1 & 53.2 & 972 & 9.000 & 24.600 & 50.9 & 59.5 & 473 & 22.289 & 34.079 & 55.8 & 65.7 & 2,588 \\
\midrule
BoT-SORT & 32.360 & 33.960 & 58.3 & 59.4 & 1,882 & 23.960 & 34.760 & 56.7 & 68.7 & 736 & 26.947 & 36.263 & 57.9 & 68.0 & 1,786 \\
BoT-SORT + CIA & 3.760 & 7.120 & 53.8 & 56.4 & 868 & 9.560 & 21.640 & 52.9 & 62.6 & 377 & 19.000 & 26.105 & 56.3 & 66.3 & 1,371 \\
\midrule
MASA & 165.440 & 165.760 & 44.9 & 39.3 & 5,707 & 56.920 & 68.840 & 56.5 & 66.9 & 1,499 & 63.421 & 69.579 & 59.8 & 71.2 & 4,256 \\
MASA + CIA & 14.840 & 18.040 & 36.4 & 37.9 & 2,013 & 10.760 & 27.400 & 51.0 & 59.9 & 729 & 16.000 & 28.211 & 58.0 & 69.8 & 3,170 \\
\bottomrule
\end{tabular}}
\end{table*}

\subsection{Final Trajectory Consolidation}
Removing final consolidation has a small effect on the three dense benchmarks in Table~\ref{tab:supp_inference_sensitivity}. Table~\ref{tab:supp_merge_additional} extends this comparison to BURST-Count and TAO-Count, where the count errors also change little. This isolates final geometric consolidation from duplicate confirmation and identity-level linking.

\section{Transfer Across Pipelines}
\label{sec:supp-transfer}

\subsection{Transfer Across Trackers}
Table~\ref{tab:supp_tracker_transfer} applies the frozen CIA module to outputs from four trackers. The experiment replays cached tracker trajectories and extracts SAM~3/IEM appearance features for CIA. It does not retrain the association network for each tracker. Each comparison uses the same input tracking output before and after CIA.

CIA reduces count error, composition error, and switches in the completed experiments. HOTA and IDF1 decrease, showing that consolidating counted identities does not automatically improve overall tracking quality. These experiments support transfer of CIA's counting decisions across trackers while revealing the cost of incorrect identity merges or suppressed observations.

\subsection{Detector and Tracker Substitution}
Table~\ref{tab:supp_pipeline_transfer} evaluates CIA with CountVid's detection-and-tracking pipeline and with CountGD-Box detections followed by SAM~3 tracking. Both use the trained CIA module without additional adaptation. These experiments change the upstream observations and tracking behavior, complementing the tracker-output comparisons above.

\begin{table*}[!h]
\centering\small
\setlength{\tabcolsep}{4pt}
\caption{Transfer to alternative detection and tracking pipelines without retraining CIA. Dance-Count uses 25 videos at 20\,fps; GMOT-Count uses all 40 videos at 30\,fps. HOTA and IDF1 are percentages.}
\label{tab:supp_pipeline_transfer}
\begin{tabular}{llrrrrr}
\toprule
Dataset & Pipeline & MAE$\downarrow$ & CE$\downarrow$ & HOTA$\uparrow$ & IDF1$\uparrow$ & ID sw.$\downarrow$ \\
\midrule
Dance-Count & CountVid + CIA & 2.880 & 4.480 & 62.5 & 64.7 & 11,156 \\
Dance-Count & CountGD-Box + SAM~3 + CIA & 3.080 & 4.360 & 65.3 & 73.0 & 1,504 \\
\midrule
GMOT-Count & CountVid + CIA & 11.025 & 18.475 & 59.6 & 70.8 & 4,746 \\
GMOT-Count & CountGD-Box + SAM~3 + CIA & 9.925 & 15.275 & 62.2 & 76.0 & 965 \\
\bottomrule
\end{tabular}
\end{table*}

The two integrations exhibit different temporal identity behavior despite competitive count errors. In particular, CountVid + CIA can reduce count error while retaining substantial identity switching. Better final counting therefore does not necessarily imply uniformly better temporal correspondence.

\section{Identity Composition and Failure Analysis}
\label{sec:suppl_animal}

\subsection{Qualitative Examples}
\begin{figure*}[!h]
\centering
\includegraphics[width=\textwidth]{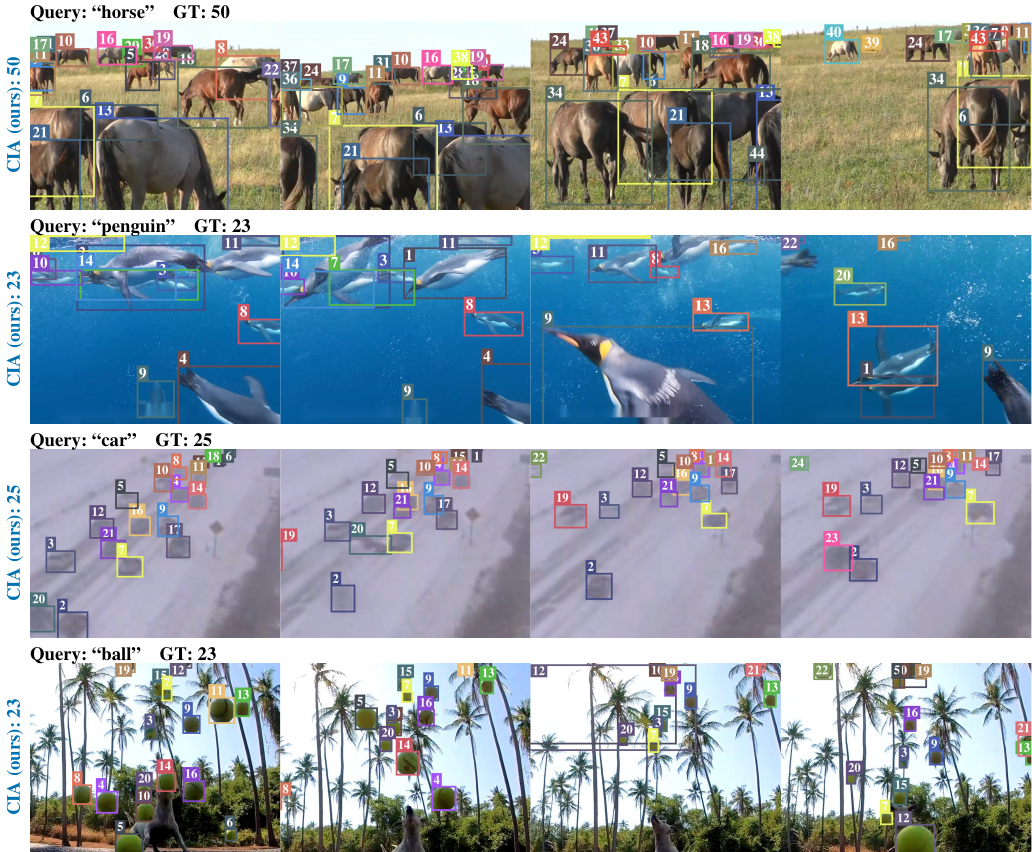}
\caption{Qualitative CIA outputs on four queries. Each block contains four sampled frames. The header gives the query and ground-truth count; the vertical label reports CIA's count. Colored boxes and numbers denote persistent identities. The examples illustrate association under motion, overlap, and partial visibility and were selected for exact or near-exact counts.}
\label{fig:supp_cia_only}
\end{figure*}

Figure~\ref{fig:supp_cia_only} shows CIA's output on four open-vocabulary queries: horse, penguin, car, and ball. The examples span crowded animals, underwater motion, small roadside objects, and a wide outdoor scene. They illustrate that CIA maintains numbered identities across four sampled frames, including through motion and partial visibility. 

\end{document}